\pdfoutput=1
\documentclass[11pt,a4paper]{article}
\usepackage[utf8]{inputenc}
\usepackage[T1]{fontenc}
\usepackage[margin=1in]{geometry} 
\usepackage{microtype}
\usepackage{amsmath,amssymb,amsfonts,mathtools,bm}
\usepackage{graphicx}
\usepackage{booktabs}
\usepackage{multirow}
\usepackage{makecell}
\usepackage{array}
\usepackage{colortbl}
\usepackage{xcolor}
\usepackage{subcaption}
\usepackage{algorithm}
\usepackage{algpseudocode}
\usepackage{enumitem}
\usepackage{xspace}

\usepackage[numbers,sort&compress]{natbib}
\usepackage[colorlinks=true]{hyperref}
\definecolor{argaccent}{HTML}{6A1B9A}
\definecolor{argdark}{HTML}{311B92}
\definecolor{ourrow}{HTML}{F2E8FF}
\definecolor{lightblue}{HTML}{EAF3FF}
\definecolor{softgreen}{HTML}{E8F5E9}
\definecolor{softred}{HTML}{FFF1F1}
\definecolor{softgray}{HTML}{F7F7F7}
\definecolor{darkgreen}{HTML}{1B5E20}
\definecolor{darkred}{HTML}{B71C1C}

\hypersetup{
  citecolor=argaccent,
  linkcolor=argaccent,
  urlcolor=argaccent
}

\newcommand{\Argus}{\textsc{Argus}\xspace}
\newcommand{\AID}{\textsc{AID}\xspace}
\newcommand{\SMII}{\textsc{SMII}\xspace}
\newcommand{\ASLG}{\textsc{ASLG}\xspace}
\newcommand{\TIA}{\textsc{TIA}\xspace}
\newcommand{\R}{\mathbb{R}}
\newcommand{\E}{\mathbb{E}}
\newcommand{\one}{\mathbf{1}}
\newcommand{\concat}{\mathbin{\Vert}}
\newcommand{\best}[1]{\textbf{#1}}
\newcommand{\second}[1]{\underline{#1}}

\newcommand{\sg}{\operatorname{sg}}

\setlist[itemize]{leftmargin=*, topsep=2pt, itemsep=1pt}
\setlist[enumerate]{leftmargin=*, topsep=2pt, itemsep=1pt}
\renewcommand{\arraystretch}{1.05}

\newcommand{\figplaceholder}[2]{%
\IfFileExists{#1}{\includegraphics[width=\linewidth]{#1}}{%
\fbox{\begin{minipage}[c][#2][c]{0.96\linewidth}
\centering
\small
\textbf{Placeholder for \texttt{#1}.}\\
Prepare this figure following the design notes in the source comments.
\end{minipage}}}}

\title{\Argus: Stacked Multi-View Identity Mosaic Injection \\ for Subject-Preserving Video Generation}

\author{%
  Zijie Meng$^{1,2}$, Jiwen Liu$^{2}$, Yufei Liu$^{3}$, Chengzhuo Tong$^{2}$, \\
  Xiaoqiang Liu$^{2}$, Yuanxing Zhang$^{2}$, Yulong Xu$^{2}$, Pengfei Wan$^{2}$ \\[1ex]
  $^1$Peking University \quad $^2$Kuaishou Technology \quad $^3$Xiamen University \\
  {\small \texttt{ymlf@stu.pku.edu.cn}}
}
\date{} %

\begin{document}
\maketitle

\begin{abstract}
Subject-preserving video generation is not solved by frontal-face similarity alone: a generated person must remain recognizable across motion, large viewpoint changes, expression shifts, occlusion, scale variation, and conflicts among text, first-frame, and identity references. We argue that the central bottleneck is the point-reference paradigm, which collapses identity into a single static observation entangled with pose, accessories, lighting, background, and camera statistics. We introduce \Argus, a Wan-based framework centered on \emph{Stacked Multi-View Identity Mosaic Injection} (\SMII). \SMII converts MLLM-selected image/video identity evidence into a \(3\times3\) stacked mosaic, synchronizes the mosaic with the current diffusion time, and injects it as negative-time read-only memory in Wan's native token space. This turns identity from an external clean adapter or a single reference image into a compact dynamic distribution. Around \SMII, an MLLM Identity Director selects informative identity moments and resolves condition conflicts, while no-cross-pair counterfactual training, Temporal Identity Annealing, and Adaptive Self-Likeness Guidance improve robustness without paired subject-video supervision. We further release HardID-Celeb, a public-figure identity-stress benchmark, and introduce YawScore and OccScore to probe large-yaw and first-frame-occlusion robustness. \Argus achieves state-of-the-art results on OpenS2V-Eval Human-Domain, reaching 64.38 Total Score, 71.86 FaceSim, 51.62 NexusScore, and 79.14 NaturalScore. On HardID-Celeb, \Argus obtains 76.80 FaceSim and improves YawScore and OccScore by 12.60 and 15.10 points over the strongest baselines, demonstrating that dynamic identity memory and large-scale counterfactual self-supervision are highly effective for subject-preserving video generation.
\end{abstract}

\section{Introduction}

Modern video foundation models have rapidly advanced through large diffusion and flow-matching transformers, enabling high-quality text-to-video and image-to-video generation~\citep{dit,sd3,flowmatching,wan21,hunyuanvideo,cogvideox,opensora,opensoraplan,stepvideo,allegro,mochi,easyanimate,magi1}. However, subject-preserving video generation remains a different and harder problem. The goal is not only to synthesize a realistic video, but to make the generated subject remain the same person while following the text prompt and respecting the first-frame condition. This requirement becomes especially demanding under large yaw, facial expression changes, partial occlusion, small faces, accessories, lighting shifts, and conflicts among the text, first frame, and identity reference.

Existing customized or identity-preserving video methods usually condition generation on one or a few reference images through adapters, token concatenation, cross-modal alignment, restricted attention, or editing-style injection~\citep{idanimator,consisid,echovideo,phantom,concatid,videomaker,ingredients,hunyuancustom,vace,skyreelsa2,magref,bindweave,standin,cinema,tpige,fantasyid}. These approaches have improved static visual similarity, but they still inherit a deeper limitation: identity is treated as a point reference. A single frame inevitably entangles who the person is with how the person happens to appear in that frame, including pose, expression, glasses, hairstyle, background, crop, and camera style. If the model over-trusts the reference, it copies nuisance details; if it under-trusts the reference, identity drifts.

\begin{figure}[ht]
\centering
\figplaceholder{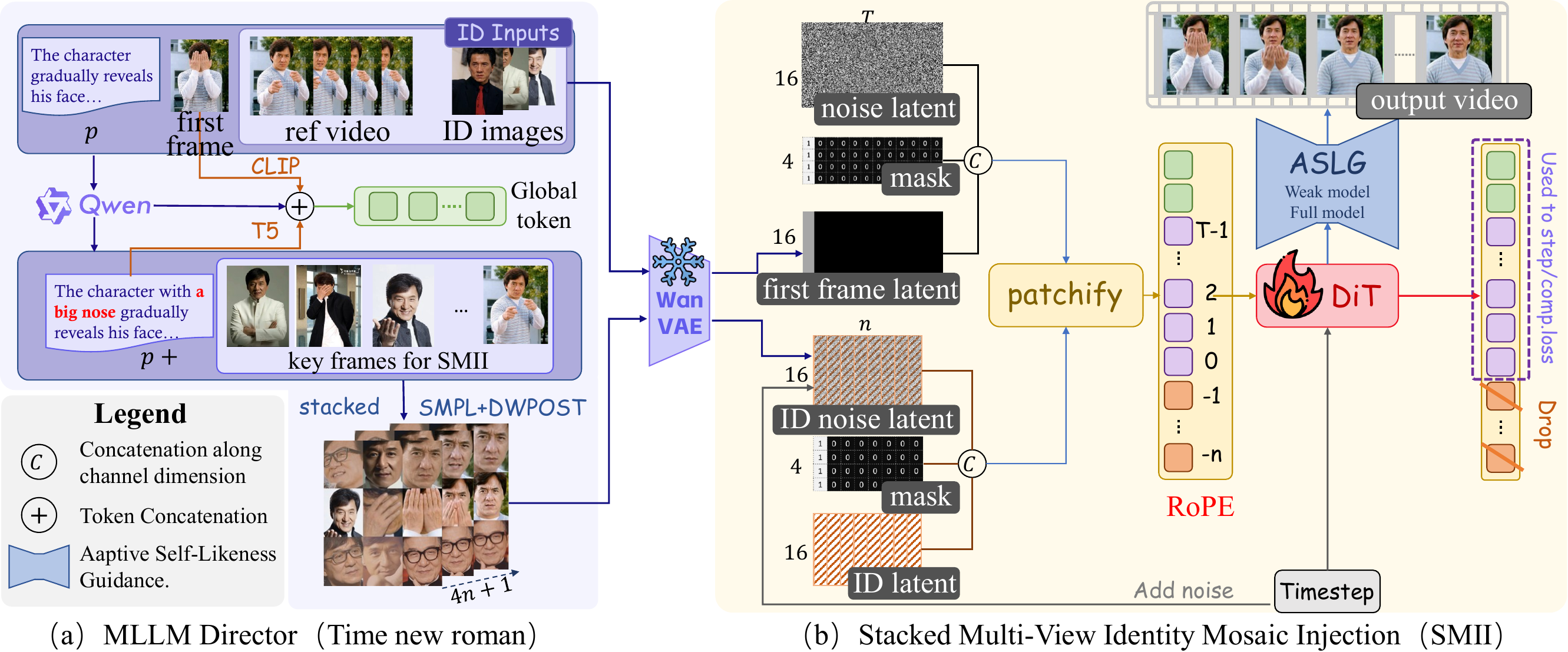}{0.25\textheight}
\vspace{-2mm}
\caption{\textbf{\Argus overview.}
A multimodal large language model (MLLM) Identity Director compiles dynamic identity evidence from reference images/videos. Stacked Multi-View Identity Mosaic Injection (\SMII) arranges selected identity observations into a $3\times3$ stacked mosaic whose cells may become $n$ latent frames after VAE compression, synchronizes the mosaic latent with the current diffusion time, and injects it as negative-time read-only memory tokens. During sampling, Adaptive Self-Likeness Guidance (\ASLG) separates text guidance from identity guidance.}
\vspace{-2mm}
\label{fig:overview}
\end{figure}

Our key observation is that human identity in video is not a point, but a distribution. People recognize a familiar subject not only from frontal geometry, but also from side profiles, facial asymmetries, habitual expressions, micro-motions, and the way appearance remains consistent across changing conditions. Thus, the target of subject-preserving generation should move beyond static resemblance toward dynamic likeness. Instead of asking the model to match one clean image, we should expose it to compact, diverse, and nuisance-randomized identity evidence.

We introduce \Argus, a Wan-based framework that turns identity references from isolated points into dynamic identity memory. \Argus accepts both images and videos as identity evidence. An MLLM Identity Director compiles informative moments from the reference bank, selects diverse identity observations, and resolves conflicts among the prompt, first frame, and reference identity. The selected evidence is then organized by Stacked Multi-View Identity Mosaic Injection (\SMII), the core module of \Argus.

\SMII represents identity as a \(3\times3\) stacked mosaic whose cells may contain either static images or short micro-clips. Rather than injecting identity through an external adapter, \SMII places the mosaic on the same diffusion path as the target video and reuses Wan's native token interface. The resulting identity tokens are assigned negative temporal positions and used as read-only memory, so the video can attend to identity evidence without contaminating the reference itself. This design gives the model a native route to read multi-view and dynamic identity cues while avoiding the distribution mismatch of clean reference-token concatenation.

A second principle of \Argus is that scalable identity supervision need not rely on cross-pair videos. Cross-pair data, where a reference and target video show the same subject under different conditions, are valuable but expensive, biased, and difficult to scale. \Argus instead uses same-clip or same-identity references with strong counterfactual augmentation: backgrounds, accessories, foreground objects, artifacts, color statistics, and crops are randomized while stable facial identity is preserved. This forces the model to ignore nuisance variables and rely on identity-invariant evidence. At inference, Temporal Identity Annealing (\TIA) and Adaptive Self-Likeness Guidance (\ASLG) separate text adherence from identity fidelity, preventing identity control from overwhelming layout early or over-sharpening faces late. In ablation, replacing this counterfactual recipe with 2K cross-pair samples reduces Total Score by 5.76 points, suggesting that large-scale nuisance randomization can be a stronger signal than raw pairing.

To evaluate this setting more faithfully, we also introduce HardID-Celeb, a public-figure benchmark designed for identity-stress scenarios such as large yaw, first-frame occlusion, small faces, accessories, and ambiguous anchors. Public figures make identity judgments more sensitive and better suited for human evaluation. In addition, we introduce two robustness metrics, YawScore and OccScore, to specifically measure identity preservation under large viewpoint changes and occluded first-frame conditions. On OpenS2V-Eval Human-Domain~\citep{opens2v}, \Argus achieves a new state of the art with 64.38 Total Score, improving over the strongest baseline by 4.18 points, while also obtaining the best FaceSim, NexusScore, and NaturalScore. On HardID-Celeb, \Argus improves FaceSim to 76.80 and achieves large gains on YawScore and OccScore, confirming that dynamic identity memory improves robustness where identity drift is most visible.

Our contributions are summarized as follows:
\begin{itemize}
    \item We propose \Argus, a subject-preserving video generation framework that accepts both image and video identity references, moving from point-reference conditioning to dynamic identity reference and from static resemblance to dynamic likeness.
    \item We introduce \SMII, a native Wan-space identity injection mechanism that represents multi-view image/video evidence as diffusion-synchronized, negative-time, read-only identity memory.
    \item We develop no-cross-pair counterfactual training and identity-aware inference, including fixed-length null mosaics, \TIA, and \ASLG, to decouple text guidance from identity fidelity without paired subject-video supervision.
    \item We release HardID-Celeb and propose YawScore and OccScore for identity-stress evaluation, and show state-of-the-art results on both OpenS2V-Eval Human-Domain and HardID-Celeb.
\end{itemize}

\section{Related Work}

\textbf{Video foundation models and customized video generation.}
Diffusion Transformers and flow-matching models have become the dominant design for large-scale video synthesis, including Wan, HunyuanVideo, CogVideoX, Step-Video, Open-Sora, Open-Sora Plan, Mochi, Allegro, EasyAnimate, and Magi-1 \citep{wan21,hunyuanvideo,cogvideox,stepvideo,opensora,opensoraplan,mochi,allegro,easyanimate,magi1}. Customized video generation extends these models with subject control, often through image adapters, temporal concatenation, face-aware features, frequency decomposition, or multimodal fusion \citep{idanimator,consisid,echovideo,phantom,concatid,videomaker,ingredients,hunyuancustom,vace,skyreelsa2,magref,bindweave,standin,fantasyid}. Earlier personalization techniques in image generation, including Textual Inversion, DreamBooth, LoRA personalization, and IP-Adapter, also inspire reference conditioning in video \citep{textualinversion,dreambooth,lora,ipadapter}. Human-centric animation and identity-specific video customization are further explored in Magic-Me and DreamDance \citep{magicme,dreamdance}. Compared with these methods, \Argus focuses on dynamic identity evidence and native diffusion-time injection rather than static reference matching.

\textbf{Identity conditioning, guidance, and evaluation.}
Commercial systems such as Vidu, Pika, Jimeng, and Hailuo demonstrate impressive video quality but offer limited transparency about identity conditioning mechanisms \citep{vidu,pika,jimeng,hailuo}. Recent academic works study MLLM-guided generation, training-free prompt/image/guidance enhancement, and multi-subject coherence \citep{cinema,tpige}. Guidance methods such as classifier-free guidance, perturbed-attention guidance, and self-guidance/autoguidance improve sampling controllability \citep{cfg,pag,autoguidance,slg}. Evaluation has expanded from generic video quality benchmarks to subject-aware and human-aligned metrics, including VBench, VBench++, EvalCrafter, FETV, T2VScore, ChronoMagic-Bench, DreamBench++, and OpenS2V-Nexus \citep{vbench,vbenchpp,evalcrafter,fetv,t2vscore,chronomagic,dreambench,opens2v}. We evaluate identity with ArcFace, CurricularFace, CLIPScore, and GME-style multimodal relevance measures \citep{arcface,curricularface,clipscore,gme}, and add stress metrics for large-yaw and occluded-anchor cases.

\section{Method}

\subsection{Preliminaries: Wan Image-to-Video Flow Matching}

Let $V\in\R^{3\times T\times H\times W}$ be the target video, $I$ the first-frame condition, $p$ the text prompt, and $R$ the identity reference bank. Wan~2.1~\citep{wan21} is a 14B-parameter DiT trained with the Flow Matching objective~\citep{flowmatching}. Its 3DVAE encodes $V$ into a clean latent $z_0^v\in\R^{16\times \widetilde T\times \widetilde H\times \widetilde W}$. Given Gaussian noise $\epsilon^v\sim\mathcal{N}(0,I)$ and noise scale $\sigma_t=t/1000$, the conditional flow sample is $z_t^v=(1-\sigma_t)z_0^v+\sigma_t\epsilon^v .$
For image-to-video conditioning, Wan composes a 36-channel tensor $\Phi_t^v=[z_t^v\concat m^v\concat y^v]$, where $m^v\in\{0,1\}^{4\times \widetilde T\times \widetilde H\times \widetilde W}$ is the temporal mask, with ones at the known first-frame condition and zeros elsewhere, and $y^v\in\R^{16\times \widetilde T\times \widetilde H\times \widetilde W}$ stores the VAE-encoded clean conditioning frame. A 3D patch embedding $\mathcal{E}_{\rm patch}$ projects $\Phi_t^v$ through a $1\times2\times2$ convolutional kernel into a token sequence consumed by stacked DiT blocks. Wan uses three independent 1D RoPE bases along the temporal, height, and width axes \citep{rope}; for head dimension $d$, the axis split is
\begin{equation}
d_f=d-2\lfloor d/3\rfloor,\qquad
d_h=d_w=\lfloor d/3\rfloor .
\end{equation}
\Argus preserves this interface and adds identity by constructing a second 36-channel stream that is distribution-aligned with $\Phi_t^v$. Appendix~\textcolor{red}{A} explains why channel-wise concatenation, cross-attention adapters, and clean prefix tokens are structurally mismatched with Wan, motivating the native diffusion-synchronized design of \SMII.

\subsection{Overview}

\Argus has three modules. The \Argus Identity Director (\AID) uses a multimodal large language model (MLLM) to select identity-informative evidence and produce conflict-aware captions. The Stacked Multi-View Identity Mosaic Injection module (\SMII) converts this evidence into a dynamic identity memory in Wan's token space. Adaptive Self-Likeness Guidance (\ASLG) calibrates identity guidance at inference. The complete pipeline is illustrated in Figure~\ref{fig:overview}. Training is deliberately simple: large-scale same-clip self-supervision, strong augmentation, and no cross-pair data.

\subsection{\Argus Identity Director}

The Identity Director is not a cosmetic captioning step; it is a perception front-end that compiles the best evidence for the downstream identity injector. We instantiate it with a frozen Qwen3-VL MLLM~\citep{qwen3vl}. Given $(R,I,p)$, the director is prompted with a hierarchical deliberation template whose complete version is provided in Appendix~\textcolor{red}{B}. It returns three outputs.

\textbf{Dynamic identity evidence.}
The director ranks candidate frames or short segments from the identity reference bank. For candidate $j$, it estimates viewpoint coverage, expression intensity, motion informativeness, scale stability, illumination diversity, visibility, and nuisance risk:
\begin{equation}
s_j=
\alpha_1 c_j^{\rm view}
+\alpha_2 c_j^{\rm expr}
+\alpha_3 c_j^{\rm motion}
+\alpha_4 c_j^{\rm scale}
+\alpha_5 c_j^{\rm illum}
+\alpha_6 c_j^{\rm visible}
-\alpha_7 c_j^{\rm nuisance}.
\end{equation}
Rather than simply choosing the top-scoring frames, we select a diverse set
\begin{equation}
\mathcal{S}^{\star}
=
\arg\max_{|\mathcal{S}|=9}
\left[
\sum_{a\in\mathcal{A}_{\rm id}}
\operatorname{Var}\big(\{c_j^a:j\in\mathcal{S}\}\big)
+\eta\sum_{j\in\mathcal{S}}s_j
-\gamma\,\operatorname{Redundancy}(\mathcal{S})
\right],
\end{equation}
where $\mathcal{A}_{\rm id}$ contains the identity-relevant axes above. This favors a reference set that spans yaw, expression, scale, illumination, and occlusion instead of over-selecting near-duplicate frontal faces. We then refine boxes and masks using whole-body pose, human motion recovery, open-vocabulary detection, and video segmentation tools \citep{dwpose,gvhmr,yoloworld,groundingdino,sam2}. Exact face-crop dilation, keypoint/vertex choices, and fallback rules are listed in Appendix~\textcolor{red}{B}. The final output is a set of nine cells, each either an image or a short micro-clip.

\textbf{Identity caption and conflict graph.}
The director produces a compact identity caption that emphasizes stable facial structure, age range, hairstyle, characteristic expression, and motion tendencies, while avoiding transient accessories. Following prompt/image/guidance enhancement practices in identity-preserving generation~\citep{tpige}, it rewrites $p$ into $p^+$ by explicitly verbalizing robust identity attributes only when they do not contradict the prompt. The director also builds a conflict graph among the first frame, text prompt, and identity bank. We enforce the priority rule: $\text{first-frame layout} \succ \text{text semantics} \succ \text{identity reference}$, so the identity bank cannot overwrite hard first-frame geometry or prompt-specified scene constraints.

\textbf{Text-side identity conditioning.}
Finally, the director emits a hidden identity summary vector $h_D$ from its final pooled hidden state. The first frame is encoded by CLIP image encoder~\citep{clip}, the rewritten prompt is encoded by T5 encoder~\citep{t5}, and $h_D$ is projected into the same conditioning width:
\begin{equation}
e_I=P_I(f_{\rm CLIP}(I)),\qquad
e_p=f_{\rm T5}(p^+),\qquad
e_D=P_D(h_D).
\end{equation}
The DiT receives the concatenated text-side conditioning
\begin{equation}
C_{\rm txt}=[e_p\concat e_I\concat e_D],
\end{equation}
with conflict-aware residual gates satisfying $g_I\ge g_p\ge g_D$. Thus, the first-frame embedding, rewritten text, and global identity descriptor jointly condition the DiT, while preserving the intended priority order.

\subsection{Stacked Multi-View Identity Mosaic Injection}
\label{sec:smii}

The core of \Argus is Stacked Multi-View Identity Mosaic Injection (\SMII), which turns identity into a time-aligned memory rather than an external adapter. \SMII combines four ideas: stacked dynamic mosaics, diffusion-synchronized identity tensorization, negative-time positional anchoring, and read-only temporally annealed attention.

\textbf{Stacked dynamic mosaic.}
Let the grid side be $g=3$, so the number of mosaic cells is $C=g^2=9$. We denote by $n$ the VAE-compressed temporal length of the identity mosaic, i.e., the number of latent identity time slots after Wan's temporal compression. In pixel space, each cell is built from either a static image or a raw micro-clip and then temporally aligned to these $n$ latent slots. For notational simplicity, we write cell $c$ as $R_c=\{r_{c,1},\ldots,r_{c,n}\}$, where a static image is repeated across all $n$ slots and a raw micro-clip is sampled or interpolated to match the VAE-aligned identity timeline. At VAE-aligned identity time $\tau$, we form a $3\times3$ canvas
\begin{equation}
G_\tau=
\operatorname{Tile}_{3\times3}
\left(
\mathcal{A}_{1,\tau}(r_{1,\tau}),
\ldots,
\mathcal{A}_{9,\tau}(r_{9,\tau})
\right),
\end{equation}
and stack these canvases into a short identity video
\begin{equation}
G_{\rm id}=(G_1,\ldots,G_n)\in\R^{3\times n\times H_{\rm id}\times W_{\rm id}}.
\end{equation}
The augmentation family $\mathcal{A}$ includes semantic background replacement, accessory perturbation, foreground-object randomization, boundary color patches, blur, compression, color jitter, synthetic skin artifacts, crop perturbation, hairstyle changes, glasses/hat/jewelry insertion, and other nuisance transformations. For high-level semantic counterfactuals, we use Nano Banana 2~\citep{nanobanana2}, with protected facial identity masks; low-level perturbations are applied online. The rule is simple: everything except the stable facial identity region may be randomized. Appendix~\textcolor{red}{C} analyzes the stacked design, RoPE geometry, grid-size choices, and raw micro-clip length, while Appendix~\textcolor{red}{D} gives the full augmentation taxonomy.

\textbf{Diffusion-synchronized identity tensorization.}
The frozen Wan VAE encodes the stacked mosaic into $z_0^{\rm id}\in\R^{16\times n\times \widetilde H_{\rm id}\times \widetilde W_{\rm id}}$. Instead of injecting clean identity tokens into a noisy video stream, we place identity on the same Flow Matching path: $z_t^{\rm id}=(1-\sigma_t)z_0^{\rm id}+\sigma_t\epsilon^{\rm id},
\qquad
\epsilon^{\rm id}\sim\mathcal{N}(0,I).$ We then build a Wan-compatible 36-channel identity tensor: $\Phi_t^{\rm id}
=
[z_t^{\rm id}\concat \one_4^{\rm id}\concat z_0^{\rm id}]$, where $\one_4^{\rm id}\in\{1\}^{4\times n\times \widetilde H_{\rm id}\times \widetilde W_{\rm id}}$. The all-one mask means every identity token is a known reference. At inference, this construction is performed at every solver step: the first 16 channels of $\Phi_t^{\rm id}$ are rebuilt as $z_t^{\rm id}$ using the current $\sigma_t$, rather than kept as the clean $z_0^{\rm id}$. This timestep-synchronized noising simulates the same noise level as the main video latent $z_t^v$, keeping the identity stream aligned with the feature prior learned by Wan's patch embedding. Crucially, $\Phi_t^{\rm id}$ has the same channel semantics as $\Phi_t^v$, so we reuse Wan's original patch embedding: $X_t^{\rm id}=\mathcal{E}_{\rm patch}(\Phi_t^{\rm id}),\qquad
X_t^v=\mathcal{E}_{\rm patch}(\Phi_t^v),\qquad
X_t=[X_t^{\rm id}\concat X_t^v].$ This eliminates the support mismatch caused by clean reference-token concatenation and avoids adding an extra cross-attention adapter.

\textbf{Negative-time identity memory.}
Identity tokens are not future video frames. We therefore assign them negative temporal coordinates:
\begin{equation}
p_f^{\rm id}\in\{-n,-n+1,\ldots,-1\},\qquad
p_f^v\in\{0,1,\ldots,\widetilde T-1\}.
\end{equation}
For $n=1$, \SMII reduces to a static identity mosaic at time $-1$. For $n>1$, identity becomes a VAE-compressed pre-video micro-history. To make identity tokens easy to read from the first frame while preventing them from being interpreted as generated content, the final identity canvas is temporally adjacent to the first video frame but remains on the negative side. Spatially, we center-align the identity mosaic with the first-frame latent grid:
\begin{equation}
p_h^{\rm id}=h+\frac{\widetilde H-\widetilde H_{\rm id}}{2},
\qquad
p_w^{\rm id}=w+\frac{\widetilde W-\widetilde W_{\rm id}}{2}.
\end{equation}
Under 3D RoPE, this gives the model a coherent geometry: identity is close enough to be attended by the first-frame and early video tokens, but positionally separated from the generated scene.

\textbf{Read-only identity attention with Temporal Identity Annealing.}
If identity queries attend to video keys, the reference memory itself becomes contaminated by generated content. We prevent this with a block attention mask:
\begin{equation}
M=
\begin{bmatrix}
0 & -\infty\\
0 & 0
\end{bmatrix},
\end{equation}
where rows are queries and columns are keys ordered as $[{\rm id},{\rm video}]$. Identity tokens attend only to identity tokens, while video tokens can read both identity and video tokens.

We further introduce Temporal Identity Annealing (\TIA), a diffusion-time and layer-dependent gate on video-to-identity attention:
\begin{equation}
\operatorname{Attn}_{\ell}(Q,K,V)
=
\operatorname{softmax}
\left(
\frac{QK^\top}{\sqrt d}+M+\log \Gamma_{\ell,t}
\right)V,
\end{equation}
where
\begin{equation}
\Gamma_{\ell,t}(q,k)=
\begin{cases}
\lambda_{\ell}(\sigma_t), & q\in{\rm video},\ k\in{\rm id},\\
1, & \text{otherwise}.
\end{cases}
\end{equation}
The default gate is a smooth middle-window schedule:
\begin{equation}
\lambda_{\ell}(\sigma_t)
=
\lambda_{\ell}^{\max}
\operatorname{sigmoid}\!\left(\frac{\sigma_{\rm hi}-\sigma_t}{\tau}\right)
\operatorname{sigmoid}\!\left(\frac{\sigma_t-\sigma_{\rm lo}}{\tau}\right).
\end{equation}
Early denoising determines global layout; late denoising polishes texture. Uniform identity injection therefore either over-constrains layout or over-sharpens faces. \TIA concentrates identity reading in the structural middle phase.

This design matches our no-cross-pair supervision. Let $i$ denote identity and $nuis$ denote nuisance variables. Same-clip references share $i$ with the target video, while counterfactual augmentation randomizes $nuis$. The training signal can be viewed as
\begin{equation}
\E_{\mathcal{A},t}
\left[
\left\|
u_\theta\!\left(z_t^v,p^+,\SMII(\mathcal{A}(R))\right)
-
(\epsilon^v-z_0^v)
\right\|_2^2
\right].
\end{equation}
Predictors relying on nuisance variables become unstable across augmented views, while identity-invariant predictors remain stable. \SMII supplies the architectural route for this invariance: identity evidence is dynamic, time-aligned, positionally separated, and read-only. Appendix~\textcolor{red}{D} formalizes this argument.

\subsection{Training and Adaptive Self-Likeness Guidance}
\label{sec:training_guidance}

\textbf{No-cross-pair training.}
For every training clip, identity cells are sampled from the same clip or same identity bank, then heavily augmented; all augmentation details are listed in Appendix~\textcolor{red}{D}. We never require a paired video of the same subject under different conditions. With probability $p_{\rm null}$, the identity mosaic is replaced by a black mosaic $G_{\varnothing}$ of identical shape. This trains a fixed-length identity-null condition, which is later used by identity guidance. The loss is the original Wan Flow Matching (FM) loss on main video tokens only:
\begin{equation}
\mathcal{L}_{\rm FM}
=
\E_{t,\epsilon}
\left[
w(t)
\left\|
\widehat u_t^v-(\epsilon^v-z_0^v)
\right\|_2^2
\right].
\end{equation}
No identity reconstruction loss, face classifier loss, or cross-pair supervision is used.

\begin{figure}[ht]
\centering
\figplaceholder{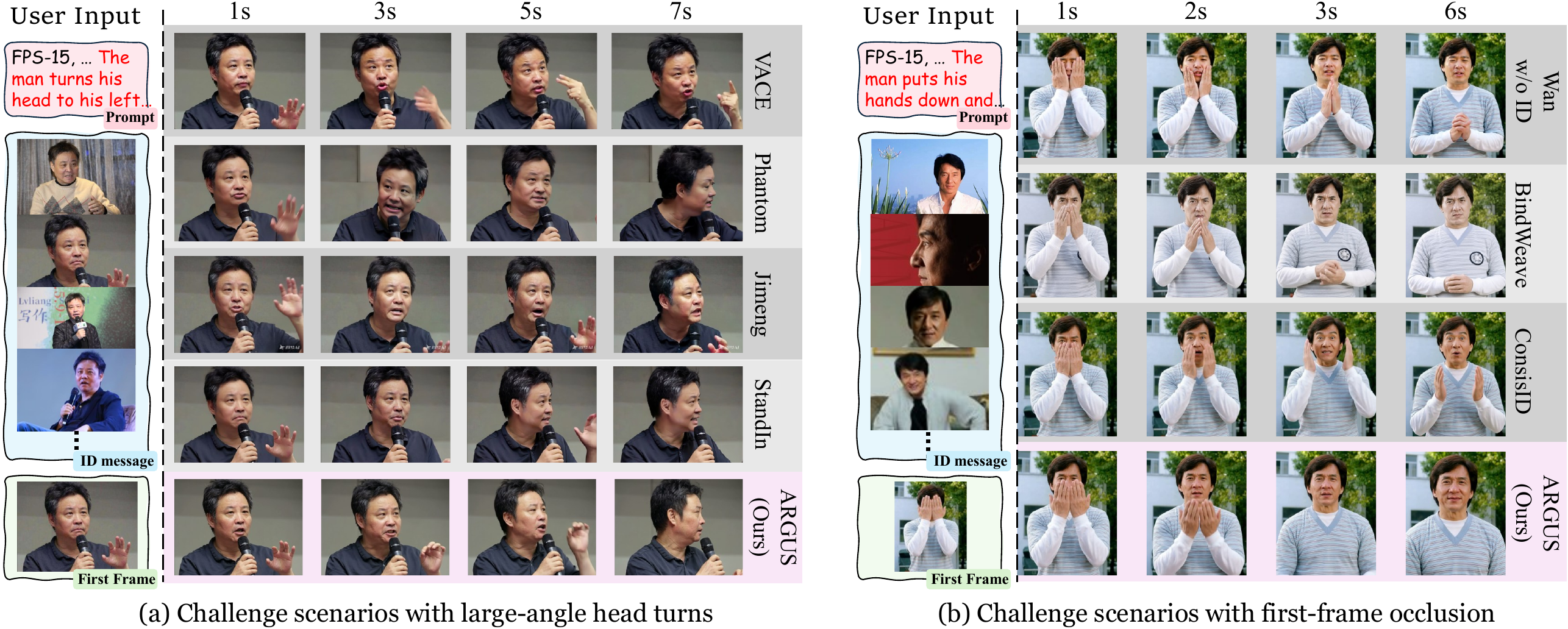}{0.25\textheight}
\vspace{-2mm}
\caption{\textbf{Qualitative comparison under identity-stress scenarios.}
(a) \textbf{Large-yaw generation.} Baselines~\citep{vace,phantom,jimeng,standin} suffer from side-face identity loss, geometric instability, or over-smoothed facial texture. \Argus preserves sharper identity details and dynamic likeness during turning.
(b) \textbf{First-frame occlusion.} Baselines either hallucinate the hidden face without identity injection~\citep{wan21}, degrade under weak first-frame cues~\citep{consisid}, or inherit copy-paste-like reference expressions~\citep{bindweave}. \Argus recovers identity from the stacked mosaic with natural expression and motion.}
\vspace{-2mm}
\label{fig:qual}
\end{figure}

\textbf{From Perturbed-Attention Guidance to Adaptive Self-Likeness Guidance.}
Standard CFG interpolates between conditional and unconditional predictions \citep{cfg}. Recent guidance methods, including Perturbed-Attention Guidance (PAG), show that constructing a weak model via attention or layer perturbation can yield a more informative direction \citep{pag,autoguidance,slg}. However, hand-designed perturbations are identity-agnostic. \ASLG learns where to weaken the model. A parameter-light mask network $\phi_\eta$, using less than 1\% of the trunk parameters, takes the timestep, global identity descriptor $h_D$, and per-block hidden statistics as input and emits a soft skip mask $\bm{\mu}=\phi_\eta(\sigma_t,h_D,\operatorname{Stat}(H))\in[0,1]^{N_{\rm blk}} .$
For DiT block $\ell$, the weak branch interpolates the residual update:
\begin{equation}
h_\ell^\omega
=
h_{\ell-1}^\omega
+
(1-\mu_\ell)\,B_\ell(h_{\ell-1}^\omega),
\end{equation}
where $B_\ell$ is the original residual block. Thus, $\mu_\ell=0$ keeps the full block and $\mu_\ell=1$ skips it. The mask network is trained with a divergence-seeking but fidelity-constrained objective based on Kullback--Leibler (KL) divergence:
\begin{equation}
\mathcal{L}_{\rm skip}
=
-
D_{\rm KL}
\left(
\mathcal{N}(u^\omega,\xi^2I)
\;\|\;
\mathcal{N}(\sg[u_{++}],\xi^2I)
\right)
+
\lambda_{\rm fid}
\left[
\frac{\|u^\omega-u_{++}\|_2}{\|u_{++}\|_2+\epsilon}
-\tau_{\rm fid}
\right]_+^2
+
\lambda_\mu\|\bm{\mu}\|_1 .
\end{equation}
The first term encourages a useful weak direction, while the constraint prevents the weak branch from collapsing into an unrelated model. The Lagrangian form and implementation details are given in Appendix~\textcolor{red}{E}.

\textbf{Adaptive Self-Likeness Guidance.}
At inference, prompt following and identity preservation should be controllable separately. We compute three velocities:
\begin{equation}
u_{++}=u_\theta(z_t,p^+,G_{\rm id}),\quad
u_{-T}=u_\theta(z_t,\varnothing,G_{\rm id}),\quad
u_{-I}=u_\theta^\omega(z_t,p^+,G_{\varnothing}).
\end{equation}
The unconditional identity branch uses the black mosaic rather than \texttt{None}, preserving sequence length and matching the $p_{\rm null}$ identity-drop training distribution. It follows the same timestep-synchronized tensorization rule as $G_{\rm id}$: at each solver step, the first 16 channels of its identity tensor are noised according to the current $\sigma_t$, so both identity-present and identity-null branches remain compatible with Wan's patch-embedding prior. The final velocity is
\begin{equation}
\widehat u_t
=
u_{++}
+
s_T(u_{++}-u_{-T})
+
s_I(\sigma_t)\,\rho_t\,(u_{++}-u_{-I}),
\end{equation}
where $s_T$ is text CFG strength, $s_I(\sigma_t)$ follows the same middle-window principle as \TIA, and
\begin{equation}
\rho_t=
\operatorname{clip}
\left(
\frac{
\|u_{++}-u_{-T}\|_2
}{
\|u_{++}-u_{-I}\|_2+\epsilon
},
\rho_{\min},\rho_{\max}
\right)
\end{equation}
self-calibrates identity guidance to the local scale of text guidance. Equivalently, \TIA can be implemented at the tensor level as
\begin{equation}
\Phi_t^{\rm id}
\leftarrow
\alpha(\sigma_t)\Phi_t^{\rm id}
+
(1-\alpha(\sigma_t))\Phi^{\varnothing},
\end{equation}
where $\alpha(\sigma_t)$ is the smooth relaxation of the hard middle window used in the attention gate. This prevents identity guidance from overwhelming layout early or over-sharpens faces late.

\section{Experiments}

\subsection{Setup}

We evaluate \Argus on OpenS2V-Eval Human-Domain~\citep{opens2v} and our HardID-Celeb benchmark. OpenS2V-Eval measures subject consistency, face similarity, motion, naturalness, and multimodal text relevance. HardID-Celeb is constructed to stress-test identity preservation under challenging human-centric conditions, including large yaw, first-frame occlusion, small faces, accessories, and ambiguous first-frame evidence. We use public figures because identity ``likeness'' is partly subjective and familiar identities enable more sensitive human judgments. All methods are evaluated with matched prompts, frame budgets, resolution, preprocessing, and metric scripts whenever applicable; full details are provided in Appendix~\textcolor{red}{F}.

\subsection{Qualitative Results and Human Study}
Figure~\ref{fig:qual} summarizes two identity-stress cases. In large-yaw generation, VACE~\citep{vace}, Phantom-14B~\citep{phantom}, Jimeng~\citep{jimeng}, and Stand-In~\citep{standin} may respectively encounter side-face drift, geometric instability, or facial over-smoothing; under first-frame occlusion, Wan~2.1~\citep{wan21} lacks explicit identity evidence, ConsisID~\citep{consisid} weakens when visible anchors are limited, and BindWeave~\citep{bindweave} can show rigid reference-like expressions. \Argus better preserves dynamic likeness by reading multi-view identity memory while \TIA and \ASLG reduce over-injection. More qualitative results are provided in Appendix~\textcolor{red}{G}, and the good/same/bad human study on HardID-Celeb is reported in Appendix~\textcolor{red}{H}.

\subsection{Quantitative Results}

\begin{figure}[ht]
\centering
\figplaceholder{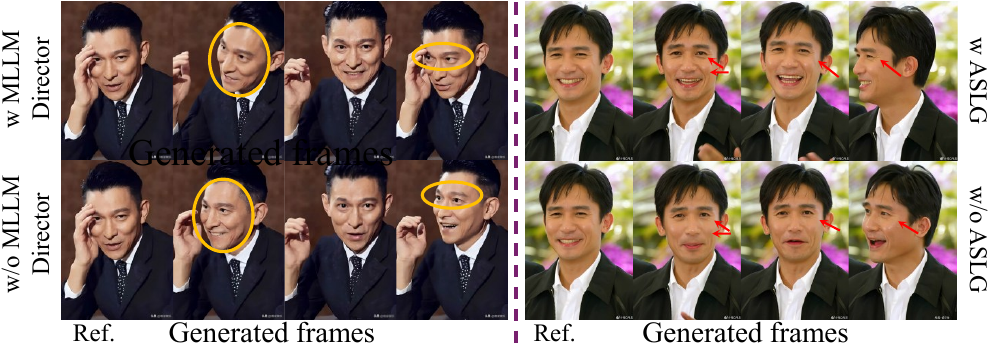}{0.25\textheight}
\vspace{-2mm}
\caption{\textbf{Ablation on the MLLM Director and \ASLG.}
The comparison shows that the MLLM Director plays a clear role in preserving fine-grained identity details, while \ASLG significantly improves identity-specific texture and skin quality.
}
\vspace{-2mm}
\label{fig:qualii}
\end{figure}

\begin{table}[ht]
\centering
\footnotesize
\renewcommand{\arraystretch}{1.05}
\setlength{\tabcolsep}{4pt}
\caption{\textbf{Quantitative comparison on OpenS2V-Eval Human-Domain.}
Top: closed-source systems. Middle: open-source methods. \textbf{Bold} = best, \underline{underline} = second-best. All methods are evaluated with identical preprocessing and metric scripts; see Appendix~\textcolor{red}{F}. \Argus is highlighted in purple. \emph{\Argus is trained without cross-pair supervision.}}
\label{tab:opens2v}
\resizebox{\linewidth}{!}{
\begin{tabular}{l|l|ccccccc}
\toprule
\rowcolor{lightblue}
\textbf{Method} & \textbf{Venue} &
\textbf{Total$\uparrow$} &
\textbf{Aesth.$\uparrow$} &
\textbf{Motion$\uparrow$} &
\textbf{FaceSim$\uparrow$} &
\textbf{GmeScore$\uparrow$} &
\textbf{NexusScore$\uparrow$} &
\textbf{NaturalScore$\uparrow$}\\
\midrule
\multicolumn{9}{l}{\textit{Closed-Source}}\\
Vidu-2.0~\citep{vidu} & API & 51.11 & 47.33 & 14.80 & 38.50 & 70.42 & 43.55 & 71.99\\
Pika-2.1~\citep{pika} & API & 52.56 & 52.39 & 28.94 & 29.41 & \second{75.03} & 45.41 & 72.53\\
Jimeng~\citep{jimeng} & API & 59.13 & 50.94 & \best{45.55} & 41.02 & 67.79 & 45.92 & \second{78.28}\\
Hailuo~\citep{hailuo} & API & \second{60.20} & 52.75 & 31.83 & 57.79 & 71.42 & 46.31 & 74.52\\
\midrule
\multicolumn{9}{l}{\textit{Open-Source}}\\
ID-Animator~\citep{idanimator} & WACV'25 & 43.37 & 42.03 & 33.54 & 31.56 & 52.91 & 35.92 & 54.03\\
ConsisID~\citep{consisid} & CVPR'25 & 52.97 & 41.76 & 38.12 & 43.14 & 72.03 & 42.08 & 64.67\\
EchoVideo~\citep{echovideo} & arXiv'25 & 54.52 & 39.93 & 35.76 & 48.57 & 68.40 & 42.96 & 69.22\\
SkyReels-A2~\citep{skyreelsa2} & arXiv'25 & 54.27 & 39.88 & 31.98 & 55.02 & 63.63 & 43.75 & 67.33\\
HunyuanCustom~\citep{hunyuancustom} & arXiv'25 & 55.85 & 49.67 & 15.13 & 62.25 & 69.78 & 44.72 & 67.00\\
MAGREF~\citep{magref} & arXiv'25 & 52.51 & 47.90 & 18.12 & 30.83 & 68.84 & 43.04 & 66.90\\
VACE-1.3B~\citep{vace} & ICCV'25 & 45.36 & 48.16 & 9.38 & 18.21 & 64.72 & 35.16 & 65.74\\
VACE-14B~\citep{vace} & ICCV'25 & 58.57 & \second{52.78} & 11.76 & 64.65 & 69.53 & 44.20 & 74.33\\
Phantom-1.3B~\citep{phantom} & arXiv'25 & 49.84 & 45.72 & 13.06 & 38.40 & 66.45 & 34.80 & 62.54\\
Phantom-14B~\citep{phantom} & arXiv'25 & 58.69 & 49.14 & 41.24 & 55.02 & 72.55 & 37.43 & 68.33\\
Stand-In~\citep{standin} & CVPR'26 & 58.36 & 50.41 & 27.18 & \second{67.31} & 72.10 & \second{47.21} & 73.05\\
BindWeave~\citep{bindweave} & ICLR'26 & 57.61 & 49.32 & 28.14 & 53.71 & 70.85 & 46.84 & 66.85\\
\midrule
\rowcolor{ourrow}
\textbf{\Argus (Ours)} & -- &
\best{64.38} & \best{54.06} & \second{42.90} &
\best{71.86} & \best{76.08} & \best{51.62} & \best{79.14}\\
\rowcolor{ourrow}
\textbf{$\Delta$ over best baseline} & -- &
\textcolor{argaccent}{\textbf{+4.18}} &
\textcolor{argaccent}{\textbf{+1.28}} &
\textcolor{black!55}{\textbf{-2.65}} &
\textcolor{argaccent}{\textbf{+4.55}} &
\textcolor{argaccent}{\textbf{+1.05}} &
\textcolor{argaccent}{\textbf{+4.41}} &
\textcolor{argaccent}{\textbf{+0.86}}\\
\bottomrule
\end{tabular}}
\vspace{-3mm}
\end{table}

Quantitatively, Tables~\ref{tab:opens2v} and~\ref{tab:hardid} show that \Argus achieves state-of-the-art overall performance while preserving naturalness: on OpenS2V-Eval Human-Domain, it improves Total Score, FaceSim, and NexusScore by 4.18, 4.55, and 4.41 points over the strongest baselines, trailing only Jimeng~\citep{jimeng} on raw Motion; on HardID-Celeb, it ranks first on FaceSim, GmeScore, NexusScore, NaturalScore, YawScore, and OccScore, and second on CLIPScore and Aesthetics. The large gains on YawScore and OccScore further confirm that \Argus recovers identity under viewpoint and occlusion stress rather than merely copying visible first-frame cues. More qualitative results and the good/same/bad human study on HardID-Celeb are in Appendix~\textcolor{red}{G} and~\textcolor{red}{H}.

\begin{table}[ht]
\centering
\footnotesize
\renewcommand{\arraystretch}{1.05}
\setlength{\tabcolsep}{3.2pt}
\caption{\textbf{Quantitative comparison on HardID-Celeb.}
Except CLIPScore, all scores are reported on a 0--100 scale after the normalization described in Appendix~\textcolor{red}{F}. GmeScore, NexusScore, NaturalScore, and Aesthetics follow the same OpenS2V-compatible scripts as Table~\ref{tab:opens2v}. The two right-most columns are robustness metrics for large yaw and occluded anchors.}
\label{tab:hardid}
\resizebox{\linewidth}{!}{
\begin{tabular}{l|cccccc|cc}
\toprule
\rowcolor{lightblue}
\textbf{Method} &
\textbf{FaceSim$\uparrow$} &
\textbf{CLIPScore$\uparrow$} &
\textbf{GmeScore$\uparrow$} &
\textbf{NexusScore$\uparrow$} &
\textbf{NaturalScore$\uparrow$} &
\textbf{Aesth.$\uparrow$} &
\textbf{YawScore$\uparrow$} &
\textbf{OccScore$\uparrow$}\\
\midrule
Hailuo~\citep{hailuo} & 52.10 & \best{21.18} & 71.06 & 46.58 & 74.80 & 52.62 & 49.80 & 46.70\\
Jimeng~\citep{jimeng} & 41.20 & 20.36 & 68.10 & 45.70 & 78.20 & \best{54.02} & 57.80 & 45.30\\
VACE-14B~\citep{vace} & 62.20 & 19.92 & 69.21 & 44.52 & 74.80 & 52.84 & 50.20 & 41.20\\
Phantom-14B~\citep{phantom} & 48.70 & 20.45 & 72.20 & 39.20 & 76.60 & 50.64 & 48.70 & 39.80\\
HunyuanCustom~\citep{hunyuancustom} & 59.80 & 19.96 & 69.70 & 44.50 & 67.40 & 50.12 & 52.10 & 42.10\\
Stand-In~\citep{standin} & \second{70.10} & 20.59 & \second{72.64} & \second{48.20} & \second{78.40} & 51.56 & \second{58.20} & \second{49.10}\\
BindWeave~\citep{bindweave} & 61.20 & 20.12 & 71.30 & 47.80 & 67.20 & 50.48 & 53.40 & 43.90\\
\midrule
\rowcolor{ourrow}
\textbf{\Argus (Ours)} &
\best{76.80} & \second{21.02} & \best{75.62} & \best{52.24} &
\best{80.62} & \second{53.60} & \best{70.80} & \best{64.20}\\
\rowcolor{ourrow}
\textbf{$\Delta$ over best baseline} &
\textcolor{argaccent}{\textbf{+6.70}} &
\textcolor{black!55}{\textbf{-0.16}} &
\textcolor{argaccent}{\textbf{+2.98}} &
\textcolor{argaccent}{\textbf{+4.04}} &
\textcolor{argaccent}{\textbf{+2.22}} &
\textcolor{black!55}{\textbf{-0.42}} &
\textcolor{argaccent}{\textbf{+12.60}} &
\textcolor{argaccent}{\textbf{+15.10}}\\
\bottomrule
\end{tabular}}
\vspace{-3mm}
\end{table}

\subsection{Ablation and Analysis}

\begin{table}[ht]
\centering
\footnotesize
\setlength{\tabcolsep}{5pt}
\renewcommand{\arraystretch}{1.05}
\caption{\textbf{Ablation on OpenS2V-Eval Human-Domain.}
Each row removes or replaces one component while keeping the rest of \Argus intact.}
\label{tab:ablation}
\resizebox{\linewidth}{!}{
\begin{tabular}{l|cccc|c}
\toprule
\rowcolor{lightblue}
\textbf{Variant} &
\textbf{Total$\uparrow$} &
\textbf{FaceSim$\uparrow$} &
\textbf{NexusScore$\uparrow$} &
\textbf{NaturalScore$\uparrow$} &
\textbf{$\Delta$Total}\\
\midrule
Wan 2.1 backbone, no ID injection & 39.21 & 12.13 & 22.41 & 51.43 & $-25.17$\\
\midrule
\Argus w/o Identity Director, random cells & 60.11 & 67.42 & 48.32 & 74.86 & $-4.27$\\
\Argus w/o text-side identity conditioning $C_{\rm txt}$ & 61.36 & 68.55 & 48.90 & 76.80 & $-3.02$\\
\Argus w/ clean identity tokens & 56.72 & 62.10 & 44.36 & 68.95 & $-7.66$\\
\Argus w/o negative-time identity positions & 58.88 & 64.32 & 46.02 & 71.64 & $-5.50$\\
\Argus w/o read-only identity memory & 59.21 & 65.18 & 46.40 & 72.33 & $-5.17$\\
\Argus w/o \TIA, uniform identity attention & 60.03 & 68.04 & 47.95 & 70.92 & $-4.35$\\
\Argus w/o \ASLG, vanilla CFG only & 61.16 & 69.12 & 48.84 & 75.61 & $-3.22$\\
\Argus replace augmentation $\to$ 2K cross-pair samples & 58.62 & 63.04 & 45.10 & 69.48 & $-5.76$\\
\Argus replace \SMII $\to$ naive token concatenation & 53.86 & 55.72 & 42.18 & 62.40 & $-10.52$\\
\midrule
\rowcolor{ourrow}
\textbf{\Argus Full} & \best{64.38} & \best{71.86} & \best{51.62} & \best{79.14} & --\\
\bottomrule
\end{tabular}}
\vspace{-3mm}
\end{table}

Table~\ref{tab:ablation} confirms that \SMII is not equivalent to naive prefixing: direct token concatenation drops Total Score by 10.52 points, and clean identity tokens without diffusion-synchronized noising drop 7.66 points. Removing negative-time positions or read-only memory also causes clear degradation, showing that identity tokens must be positionally separated from generated video tokens and protected from contamination. The Identity Director, text-side identity conditioning, \TIA, and \ASLG provide complementary gains, with qualitative comparisons shown in Figure~\ref{fig:qualii}. Replacing counterfactual same-identity training with 2K cross-pair samples reduces Total Score by 5.76 points. Additional mosaic geometry and raw-length ablations are provided in Appendix~\textcolor{red}{C}.

\section{Conclusion}

\Argus reframes subject-preserving video generation from point-reference matching to dynamic identity-memory construction. By combining \SMII with no-cross-pair counterfactual training and identity-aware guidance, \Argus achieves state-of-the-art identity consistency and robustness on OpenS2V-Eval Human-Domain and HardID-Celeb, validating dynamic identity evidence as an effective alternative to static reference conditioning.

\clearpage
\bibliographystyle{plainnat}
\bibliography{mybib}

\end{document}